\documentclass[]{spie}  

\usepackage{amsmath,amsfonts,amssymb}
\usepackage{graphicx}
\usepackage{subcaption}
\usepackage[colorlinks=true, allcolors=blue]{hyperref}

\title{Weakly-Supervised Lesion Segmentation on CT Scans \\ using Co-Segmentation}

\author[ ${\star}$${\dagger}$a]{Vatsal Agarwal}
\author[ ${\star}$a]{Youbao Tang}
\author[ b]{Jing Xiao}
\author[ a]{Ronald M. Summers}
\affil[a]{Imaging Biomarkers and Computer-Aided Diagnosis Laboratory, Radiology and Imaging Sciences, National Institutes of Health Clinical Center, Bethesda, MD 20892, USA}
\affil[b]{Ping An Insurance Company of China, Shenzhen, 510852, China}

\authorinfo{$^{\star}$ equal contribution, $^{\dagger}$ work done during internship in NIH while pursuing undergraduate degree at University of Maryland.}

\begin{document} 
\maketitle
\vspace*{-0.5\baselineskip}
\begin{abstract}
 Lesion segmentation on computed tomography (CT) scans is an important step for precisely monitoring changes in lesion/tumor growth. This task, however, is very challenging since manual segmentation is prohibitively time-consuming,  expensive, and requires professional knowledge. Current practices rely on an imprecise substitute called response evaluation criteria in solid tumors (RECIST). Although these markers lack detailed information about the lesion regions, they are commonly found in hospitals' picture archiving and communication systems (PACS). Thus, these markers have the potential to serve as a powerful source of weak-supervision for 2D lesion segmentation. To approach this problem, this paper proposes a convolutional neural network (CNN) based weakly-supervised lesion segmentation method, which first generates the initial lesion masks from the RECIST measurements and then utilizes co-segmentation to leverage lesion similarities and refine the initial masks. In this work, an attention-based co-segmentation model is adopted due to its ability to learn more discriminative features from a pair of images. Experimental results on the NIH DeepLesion dataset demonstrate that the proposed co-segmentation approach significantly improves lesion segmentation performance, \textit{e.g} the Dice score increases about 4.0\% (from 85.8\% to 89.8\%).
\end{abstract}


\keywords{Weakly-supervised lesion segmentation, segmentation, co-segmentation, attention, CT scans}

\section{Introduction}
\label{sec:purpose}  

As clinical protocols use response evaluation criteria in solid tumors (RECIST) in computed tomography (CT) images for cancer patient monitoring, many hospitals' picture archiving and communication systems (PACS) store a vast number of lesion diameter measurements paired with corresponding CT images. These markers are determined by a radiologist who for each lesion, selects an axial slice where it is the largest. The diameters of the lesion are then determined for both the major and minor axis. 

RECIST markers are imperfect as the measurements are relatively subjective and can be flawed due to variation in slices or between different observers. Such inconsistencies are problematic as it is more difficult to properly measure tumor growth over a period of time, thereby impeding treatment options for patients. With a precise measurement, follow-up and quantitative analysis of tumor extents could be performed accurately to provide valuable information for treatment planning and tracking. However, performing full volumetric lesion measurements are highly costly and time-consuming. This establishes the need for computational methods to automate such a task. 

Furthermore, there are significant differences between lesions with respect to size, shape and appearance which make segmentation more difficult. To address this issue, we employ a co-segmentation architecture that can extract joint semantic information from a pair of CT scans to achieve a higher accuracy. In order to ensure that the pair of lesions are similar with respect to appearance and background, we cluster the lesions into 200 groups, from which pairs are then extracted. Thus, we leverage existing RECIST diameters as weak supervision for a convolutional neural network based weakly-supervised segmentation method that employs an attention-based co-segmentation method to obtain refined lesion masks. To the best of our knowledge, this is the first work to utilize the co-segmentation strategy for lesion segmentation on CT scans.

The rest of the paper is organized as follows. In Section~\ref{sec:relatedworks}, we discuss the related works. We then present the co-segmentation architecture and training methodology in Section~\ref{sec:methods}. Lastly, we examine the quantitative and qualitative results in Section~\ref{sec:results}.  

\section{Related Works}
\label{sec:relatedworks}


Recently, many useful and important applications in medical image analysis have been developed using deep learning, such as measurement estimation \cite{tang2018semi}, lung segmentation \cite{jin2018ct,tang2019xlsor}, lesion detection \cite{yan2019mulan,tang2019lesiondetection} and segmentation \cite{cai2018accurate,tang2018ct,isbi2020}, lymph node segmentation \cite{tang2019ct}, disease classification\cite{tang2019tuna,8759442,tang2019deep}, etc.
Despite the recent success of convolutional neural network methods for different tasks, they are flawed by the requirement of needing extensive amounts of highly-annotated training data, especially for segmentation tasks. Weakly-supervised learning methods avoid this issue by utilizing weak annotations for training such as bounding boxes for the objects of interest or scribbles over the foreground\cite{khoreva2017simple}. These are then used to produce initial labels that a CNN model can then use to train and improve upon. This presents the need to ensure that proper labels can be extrapolated from weak annotations. Many popular techniques exist for this task such as GrabCut\cite{rother2004grabcut} and MCG\cite{pont2016multiscale}. Previous work\cite{cai2018accurate} exploring weakly-supervised segmentation for lesion segmentation applied GrabCut to generate the initial masks from the RECIST-slices which acted as weak supervision.   


Object co-segmentation is the task of jointly segmenting common objects from a pair of images. Proposed by Ref~\citenum{vicente2011object}, this method was shown to achieve better performance compared to segmenting objects from each image independently. While many works utilized hand-crafted features such as color histograms, SIFT descriptors and HOG features, these were not enough to handle more difficult cases where there is greater variation in the images. Recently, new techniques have been developed that employ deep learning to improve accuracy in comparison to these methods. Ref~\citenum{mukherjee2018object} first developed a deep learning strategy which used a Siamese network to obtain feature vectors for each image. They then used the ANNOY library to measure the distance between the vectors and generated a collage of masks using a fully-convolutional network to segment similar object proposals. This was then improved upon by Ref~\citenum{li2018deep} which developed a Siamese encoder-decoder network that applied a mutual correlation layer to capture and emphasize similarities between the image features. Recently, Ref~\citenum{chen2018semantic} modified this architecture by replacing the correlation layer with channel-wise and spatial-wise attention to enhance common features and suppress varying ones. 

\section{METHODS}
\label{sec:methods}
\begin{figure} [ht]
   \begin{center}
   \begin{tabular}{c} 
   \includegraphics[height=7.2cm]{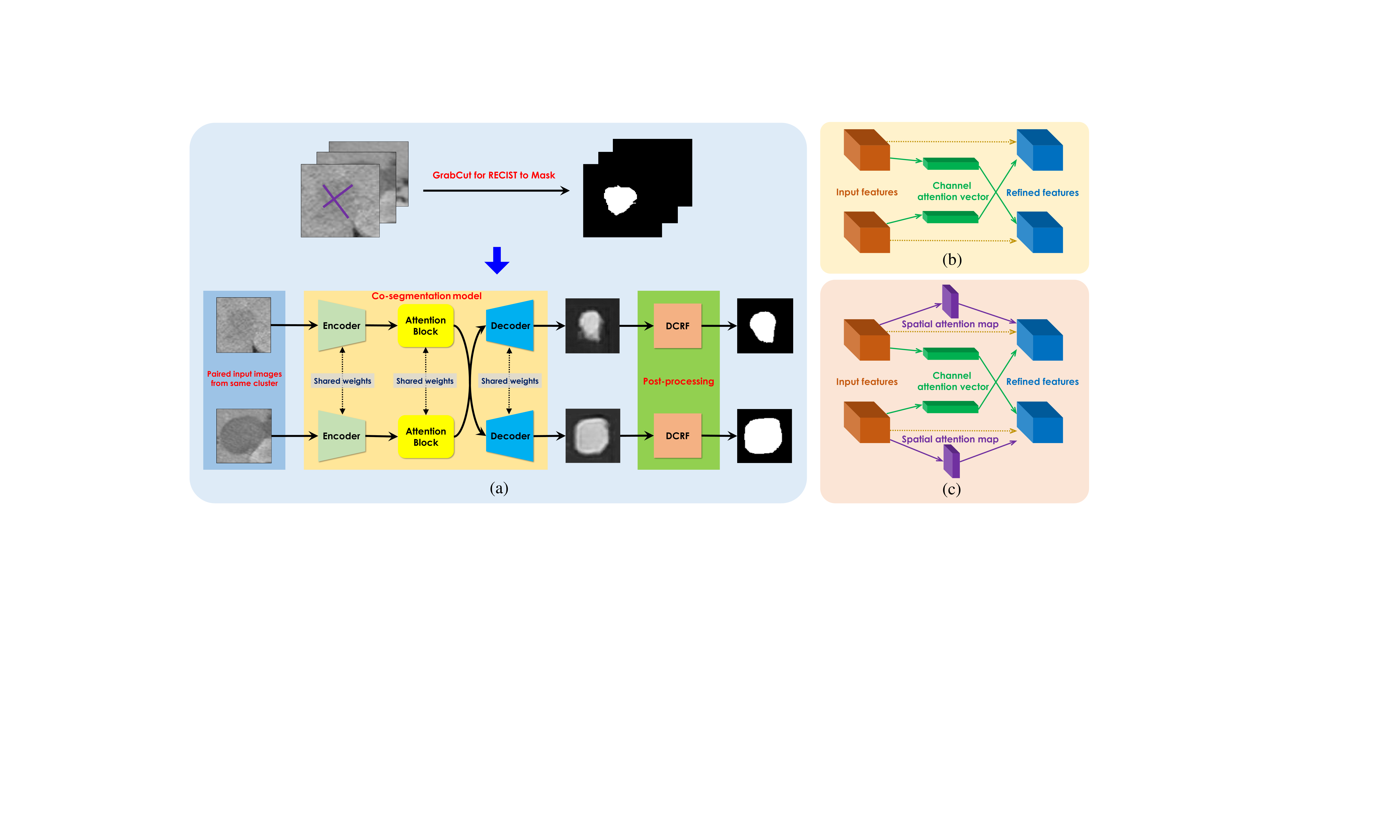}
   \end{tabular}
   \end{center}
   \caption[model-pipeline] 
   { \label{fig:model-pipeline} 
The overall pipeline of our proposed method (a) and two attention mechanisms (\textit{i.e.} channel-wise (b) and channel-spatial-wise (c)). First, GrabCut is used to create the initial lesion masks from RECIST measurements for training the co-segmentation model. The co-segmentation model takes a pair of lesion images and uses an encoder-decoder framework with an attention mechanism to enhance shared semantic information. The final segmentations are then refined by a densely-connected conditional random field (DCRF).}
   \end{figure}

In this work, we present a weakly-supervised segmentation approach to generate improved 2D lesion segmentations. Fig.~\ref{fig:model-pipeline} shows our two-tiered approach, \textit{i.e.} we first generate the initial lesion masks from RECIST measurements as ground-truth data to then train a robust attention-based co-segmentation model to produce the final refined segmentations. RECIST diameters (indicated by the purple cross in Fig.~\ref{fig:model-pipeline}(a)) act as the weakly-supervised training data. The details are described below. 

\subsection{Initial Lesion Mask Generation}
\label{sec:init slice seg}
GrabCut\cite{rother2004grabcut} is a popular method that takes in seeds for the image foreground and background regions and minimizes an objective energy function to generate segmentations. Following Ref.~\citenum{cai2018accurate}, we utilize GrabCut to leverage RECIST-slice information and compute the seeds to obtain rough lesion masks from the RECIST annotations (as shown in the top part of Fig.~\ref{fig:model-pipeline}(a)). We then use these masks as training data for the co-segmentation model. 

\subsection{Lesion Co-Segmentation}
\label{sec:lesion co-segmentation}
With the initial lesion masks, we train a co-segmentation model for joint lesion segmentation. The bottom part of Fig.~\ref{fig:model-pipeline}(a) shows a diagram of how we train the co-segmentation model using the initial lesion masks. The co-segmentation model is adopted from Ref.~\citenum{chen2018semantic} and consists of a Siamese encoder-decoder framework coupled with an attention module. The model takes in a pair of images as input and produces segmentations for each lesion, which are then passed to a densely-connected conditional random field \cite{krahenbuhl2011efficient} (DCRF) to obtain the final lesion masks.  
  
The attention mechanism consists of a channel-attention module and a spatial-attention model. Fig.~\ref{fig:model-pipeline}(b) shows only the channel attention mechanism and Fig.~\ref{fig:model-pipeline}(c) shows the channel-spatial attention mechanism (CSA). Feature maps obtained from the encoder network are passed to the mechanism to preserve shared semantic information and suppress the rest. The channel-attention module is inspired by the Squeeze and Excitation architecture\cite{hu2018squeeze} and enhances common information by retaining channels that have high activations in both images and the spatial-attention module captures inter-spatial information in the feature maps. Please refer to Ref.~\citenum{chen2018semantic} for details about the respective architectures of these attention modules. 

For the encoder network, we experiment with a variety of CNN architectures, such as VGG-16\cite{simonyan2014very} and ResNet-101\cite{he2016deep} for feature extraction. Taking inspiration from DeepLabV2\cite{chen2017deeplab}, our final model's encoder (denoted DRN-101) is a pre-trained residual network that utilizes atrous convolutions for the last two residual blocks, creating a feature map with an output stride of 8. This helps in obtaining higher-resolution feature representations with greater context. The model utilizes the channel-spatial attention module to retain more spatial information. The decoder network employs up-sampling and produces a two-channel result that represents the foreground and background predictions. 

\section{Results}
\label{sec:results}
Our model is trained on the NIH DeepLesion\cite{yan2018deeplesion} dataset which contains 32,735 lesion images with RECIST annotations. Since there is large variance between different lesions in size and appearance, we utilize the strategy from Ref.~\citenum{yan2018deep} and obtain feature vectors for each slice. K-means clustering is then applied to group the lesions into 200 classes empirically. We then use stratified sampling based on the initial clustering to split the dataset into 80\% training, 10\% validation and 10\% testing sets. A co-segmentation dataset is created by pairing images from each cluster. This results in 270,470 pairs for training, 28,136 pairs for validation and 3,866 pairs for testing. We use 1,000 manually annotated segmentations for evaluation and report the recall, precision, Dice similarity coefficient, averaged Hausdorff distance (AVD), and volumetric similarity (VS) for quantitative metrics, which are calculated pixel-wisely by a publicly available segmentation evaluation tool\cite{taha2015metrics}.
\textbf{Pre-processing:} Prior to training, we pad lesion-mask images before resizing them to 128x128 and then normalize them.
\textbf{Training:} We adopt the original network's hyper-parameters, utilizing the Adam optimizer with a learning rate of $1\mathrm{e}{-5}$ and a weight decay of 0.0005 for L2 regularization. For the loss function, we use pixel-wise cross-entropy. We use a batch-size of 20 and train the model for two epochs with 12,000 iterations per epoch. For inference, we use the model that performs the best on the validation set with regards to Dice score. 

\begin{table}[ht]
\caption{The performance of lesion segmentation with different training strategies in terms of recall, precision, Dice score, AVD, and VS. The means and standard deviations are reported for all strategies. $\uparrow$: the larger the better. $\downarrow$: the smaller the better.} 
\label{tab:Co-Segmentation-Performance}
\footnotesize
\begin{center}       
\begin{tabular}{| c | c | c | c | c | c |}
\hline
\rule[-1ex]{0pt}{3.5ex} \textbf{Model} & \textbf{Recall}$\uparrow$  & \textbf{Precision}$\uparrow$  & \textbf{Dice}$\uparrow$  & \textbf{AVD}$\downarrow$ & \textbf{VS}$\uparrow$   \\
\hline
\rule[-1ex]{0pt}{3.5ex}  FCN-16 & 
0.893 $\pm$ 0.13  & 0.841 $\pm$ 0.13  & 0.858 $\pm$ 0.11  & 0.521 $\pm$ 1.42  & 0.924 $\pm$ 0.08  \\
\hline
\rule[-1ex]{0pt}{3.5ex}  VGG-16\textsubscript{without clustering} & 
0.800 $\pm$ 0.13  & 0.844 $\pm$ 0.11  & 0.811 $\pm$ 0.09  & 0.683 $\pm$ 1.08  & 0.908 $\pm$ 0.08  \\
\hline
\rule[-1ex]{0pt}{3.5ex}  VGG-16 & 
0.870 $\pm$ 0.13  & 0.898 $\pm$ 0.09  & 0.877 $\pm$ 0.10  & 0.403 $\pm$ 1.19  & 0.935 $\pm$ 0.08  \\
\hline
\rule[-1ex]{0pt}{3.5ex}  VGG-16 + CSA & 
 0.880 $\pm$ 0.12  & 0.895 $\pm$ 0.10  & 0.880 $\pm$ 0.10  & 0.393 $\pm$ 1.22  & 0.938 $\pm$ 0.08  \\
\hline
\rule[-1ex]{0pt}{3.5ex}  ResNet-101 & 
0.857 $\pm$ 0.14  & 0.920 $\pm$ 0.09  & 0.879 $\pm$ 0.10  & 0.420 $\pm$ 1.49  & 0.928 $\pm$ 0.09  \\
\hline 
\rule[-1ex]{0pt}{3.5ex}  ResNet-101 + CSA & 
0.862 $\pm$ 0.13  & 0.921 $\pm$ 0.09  & 0.883 $\pm$ 0.10  & 0.385 $\pm$ 1.24  & 0.931 $\pm$ 0.08  \\
\hline 
\rule[-1ex]{0pt}{3.5ex}  DRN-101 & 
0.851 $\pm$ 0.16  & \textbf{0.932 $\pm$ 0.08}  & 0.877 $\pm$ 0.13  & 0.490 $\pm$ 1.39  & 0.920 $\pm$ 0.12  \\
\hline 
\rule[-1ex]{0pt}{3.5ex}  DRN-101 + CSA & 
\textbf{0.915 $\pm$ 0.11}  & 0.895 $\pm$ 0.10  & \textbf{0.898 $\pm$ 0.10} & \textbf{0.349 $\pm$ 1.20}  & \textbf{0.942 $\pm$ 0.08} \\
\hline

\end{tabular}
\end{center}
\end{table}

\subsection{The Performance of Lesion Co-Segmentation}
\label{sec: lesion co-segmentation results}

To validate the efficacy of using the co-segmentation model, we first train a fully-convolutional network (FCN)\cite{long2015fully} with a VGG-16 backbone on the initial lesion masks. For the co-segmentation models, we experiment with the VGG-16, ResNet-101 architectures as well as a dilated ResNet-101 model. The default attention mechanism is channel attention, and we apply the same DCRF model for all experiments to obtain the final segmentations. All models are implemented in PyTorch\cite{paszke2017automatic}. 

Quantitative results of these experiments are shown in Table.~\ref{tab:Co-Segmentation-Performance}. Given a 6.5\% increase in Dice score, we can see that clustering lesions prior to training greatly improves performance, supporting our methodology of leveraging lesion embeddings to generate lesion classes for training. The results also validate the point that co-segmentation produces better segmentations as the Dice score improves by 1.9\% between the FCN model and the VGG-16 co-segmentation model with only channel-wise attention. When we adopt a deeper network ResNet-101, the model gains minor improvements, achieving an 87.9\% Dice score. It is worth noting the increase in recall when using channel-spatial attention compared to channel attention. This is especially significant when examining the difference between the DRN-101 and DRN-101 + CSA masks. Thus, it can be seen that adding spatial attention considerably improves performance by reducing the number of pixels incorrectly labeled as part of the background. Additionally, the results show that adding the atrous convolutions improves the precision of the masks with a 1.2\% improvement between the ResNet-101 backbone and the DRN-101 backbone models. Our best model uses the convolutional layers of a dilated ResNet-101 with CSA (DRN-101 + CSA) for the encoder and gains a 4.0\% increase in Dice score compared to the FCN model.

\begin{figure} [ht]
    \begin{center}
        \begin{tabular}{c} 
            \includegraphics[height=8.8cm]{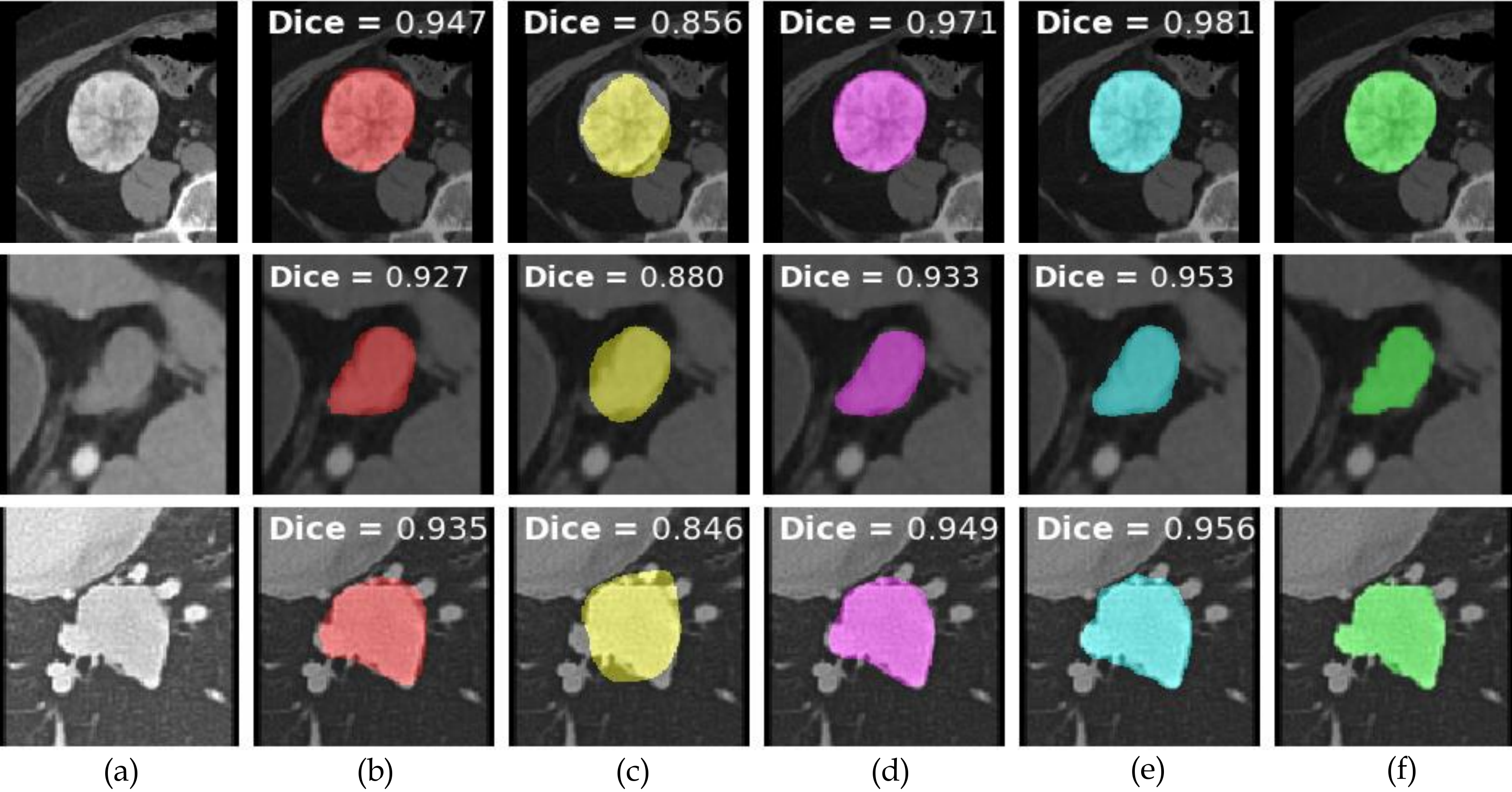}
        \end{tabular}
   \end{center}
\caption[results] 
{ \label{fig:results} Three lesion segmentation results using (b) FCN-16, (c) VGG-16 without clustering, (d) VGG-16, (e) DRN-101 + CSA. (a) is the input image, while (f) is the corresponding ground truth.}
\end{figure}

Qualitative results are shown in Fig.~\ref{fig:results}. Firstly, it can be seen that the co-segmentation model trained on the dataset without clustering suffers greatly in performance as it produces masks that are unable to capture the fine boundaries of the lesion. Generally, the model is prone to falsely classifying pixels as part of the background. This is most likely due to the variations in texture and shade between different lesions, which confuse the model. Following clustering using the lesion embeddings, it can be seen that the masks produced by the VGG-16 backbone model have less false positives compared to those produced by the FCN model. This would indicate that using semantic information from both images aids in generating more precise boundaries. Further improvements in recall can be seen between the masks created by the DRN-101 model with channel-spatial attention as less pixels are falsely labeled as the background. 

\section{CONCLUSIONS}
\label{sec:conclusions}
In this work, a weakly-supervised attention-based deep co-segmentation approach is proposed to acquire more accurate lesion segmentations from RECIST measurements. Given a set of RECIST diameters, we first utilize GrabCut to obtain initial lesion masks. We then leverage lesion embeddings to create lesion clusters, which serve as different classes for the co-segmentation model. We experiment with various architectures and attention mechanisms to demonstrate the efficacy of our approach. The quantitative and qualitative experimental results demonstrate that co-segmentation with clustering enhances lesion segmentation performance. Future work can then employ this model architecture in a slice-propagated fashion to obtain a fully volumetric lesion segmentations.

\acknowledgments{This research was supported by the Intramural Research Program of the National Institutes of Health Clinical Center and by the Ping An Insurance Company through a Cooperative Research and Development Agreement. We thank Nvidia for GPU card donation.}

\bibliography{report} 
\bibliographystyle{spiebib} 

\end{document}